\documentclass[journal]{IEEEtran}
\usepackage{amsmath,amsfonts}
\usepackage{algorithmic}
\usepackage{algorithm}
\usepackage{array}
\usepackage[caption=false,font=normalsize,labelfont=sf,textfont=sf]{subfig}
\usepackage{textcomp}
\usepackage{stfloats}

\usepackage{url}
\usepackage{verbatim}
\usepackage{mathrsfs}
\usepackage{cite}
\usepackage{graphicx}
\usepackage{bbding}

\usepackage{xcolor, colortbl}
\usepackage{times}
\usepackage{epsfig}
\usepackage{graphicx}
\usepackage{amsmath}
\usepackage{amssymb}
\usepackage{bm}
\usepackage{booktabs}
\usepackage{setspace}
\usepackage{float}
\usepackage{color}
\usepackage{multirow}

\definecolor{lightgray}{gray}{0.85}
\usepackage[colorlinks=true,citecolor=green]{hyperref}
\newcommand{\figref}[1]{Fig.~\ref{#1}}%
\newcommand{\tabref}[1]{Table~\ref{#1}}%

\renewcommand{\eqref}[1]{Eq.~(\ref{#1})}

\newcommand{\etal}{\textit{et al.}}

\newcounter{sectionCounter}
\usepackage{tabularx}

\begin{document}

\title{Multimodal Large Models Are Effective\\ Action Anticipators}
\author{Binglu Wang,~\IEEEmembership{Member,~IEEE,} Yao Tian, Shunzhou Wang,~\IEEEmembership{Member,~IEEE,} and Le Yang
\thanks{This work was supported in part by the China Postdoctoral Science Foundation under Grant 2023TQ0344, 2023M733387; the National Natural Science Foundation of China under Grant 62401447, 62306101; the State Energy Key Laboratory for Carbonate Oil and Gas under Grant 33550000-24-ZC0613-0101; and the Key Research and Development Program of Shaanxi (Program No.2024GX-YBXM-051, 2024CY2-GJHX-10). (\textit{Corresponding author: Le Yang}.)}

\thanks{Binglu Wang, Yao Tian and Shunzhou Wang are with College of Information and Control Engineering, Xi’an University of Architecture and Technology, Xi’an 710055, PR China. Binglu Wang is also with the School of Astronautics, Northwestern Polytechnical University, Xi'an 710072, China. Shunzhou Wang is also with School of Electronic and Computer Engineering, Shenzhen Graduate School, Peking University, Shenzhen 518055, China (e-mail: \{wbl19921129, 2tianyao1\}@gmail.com, shunzhouwang@163.com).} 

\thanks{Le Yang is with Institute of Artificial Intelligence, Hefei Comprehensive National Science Center. Le Yang is also with the School of Information Science and Technology, University of Science and Technology of China, Hefei, 230022,  China (e-mail: nwpuyangle@gmail.com).}

}

\markboth{IEEE Transactions on Multimedia}%
{Shell \MakeLowercase{\textit{et al.}}: A Sample Article Using IEEEtran.cls for IEEE Journals}

\IEEEpubid{}

\maketitle


\begin{abstract}
The task of long-term action anticipation demands solutions that can effectively model temporal dynamics over extended periods while deeply understanding the inherent semantics of actions. Traditional approaches, which primarily rely on recurrent units or Transformer layers to capture long-term dependencies, often fall short in addressing these challenges. Large Language Models (LLMs), with their robust sequential modeling capabilities and extensive commonsense knowledge, present new opportunities for long-term action anticipation. In this work, we introduce the ActionLLM framework, a novel approach that treats video sequences as successive tokens, leveraging LLMs to anticipate future actions. Our baseline model simplifies the LLM architecture by setting future tokens, incorporating an action tuning module, and reducing the textual decoder layer to a linear layer, enabling straightforward action prediction without the need for complex instructions or redundant descriptions. To further harness the commonsense reasoning of LLMs, we predict action categories for observed frames and use sequential textual clues to guide semantic understanding. In addition, we introduce a Cross-Modality Interaction Block, designed to explore the specificity within each modality and capture interactions between vision and textual modalities, thereby enhancing multimodal tuning. Extensive experiments on benchmark datasets demonstrate the superiority of the proposed ActionLLM framework, encouraging a promising direction to explore LLMs in the context of action anticipation. Code is available at \href{https://github.com/2tianyao1/ActionLLM.git}{\textit{https://github.com/2tianyao1/ActionLLM.git}}.

\end{abstract}


\section{Introduction}
Long-term action anticipation is essential for machines to deeply decipher patterns of human behavior, advancing applications in augmented reality, virtual reality, intelligent surveillance systems, and human-computer interaction. However, long-term action anticipation faces two primary challenges: First, it requires the model to possess strong long-term sequence modeling capabilities to handle extended future prediction time spans and the inherent uncertainty, complicating the task of capturing dependencies between distant actions. Second, the model needs to understand the intrinsic meaning of actions and discern the relationships among them. To improve comprehension, the model should also have reasoning abilities that go beyond simple correlations.

\begin{figure}[t]
\centerline{\includegraphics[width=\linewidth]{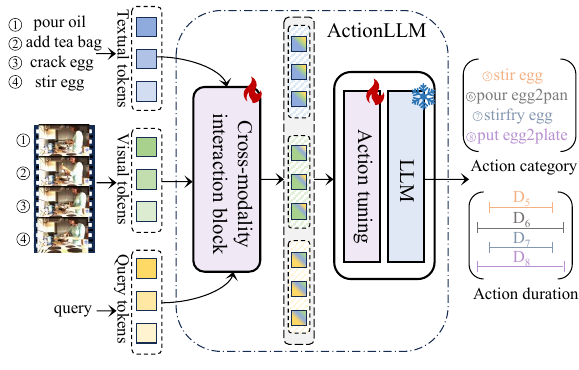}}
\captionsetup{justification=raggedright,singlelinecheck=false}
\caption{An application of ActionLLM for predicting long-term actions in kitchen scenarios. We leverage the LLM to explore the interdependencies among these actions. By integrating text labels with visual features, synchronized with frames, we achieve information fusion through a Cross-Modality Interaction Block. The text aligns with the LLM's input format, enhancing the utilization of its intrinsic commonsense for precise prediction of future action categories and their durations (D).}
\vspace{-0.4cm}
\label{example}
\end{figure}

A considerable body of research seeks to handle the challenges associated with long-term action anticipation~\cite{abu2021long,moniruzzaman2022jointly} and has produced notable outcomes~\cite{zhang2024object,patsch2024long,abu2019uncertainty}. Nevertheless, traditional LSTM and RNN-based approaches~\cite{abu2018will,ke2019time,sener2020temporal} encounter difficulties in capturing long-range dependencies, exhibit low computational efficiency, and struggle to generalize to novel scenarios. Recently, Transformer-based methods~\cite{gong2022future,nawhal2022rethinking,bhagat2024let,wang2023temporal} excel in processing sequential data; however, they are constrained by low parameter efficiency, which increases with sequence length. Additionally, their sensitivity to input sequences can result in performance variability. The advent of Large Language Models (LLMs)\cite{chowdhery2023palm,radford2019language,touvron2023llama} presents new possibilities for action anticipation. LLMs accumulate commonsense knowledge from extensive textual data, providing a strong foundation for understanding the meaning of actions. Moreover, the performance of LLMs in sequential perception tasks\cite{wang2023timemixer,gruver2024large} reflects their advanced capabilities in long-term sequence modeling, which is crucial for predicting long-term actions.

Thus, as a preliminary exploration, this study directly handles the long-term action anticipation task using LLMs. Instead of utilizing intricate instructions or superfluous descriptions to bridge the gap between visual data and language comprehension mechanisms, we adopt an approach that prioritizes intuitive integration. 
On the input side, we implement a feature adapter strategy, designed to seamlessly amalgamate visual features within the framework of the LLM.
On the output side, the output mechanism of the LLM is streamlined by replacing the original text decoding layer with a linear layer, thereby enabling the language model to present long-term action anticipation results in a more straightforward manner. 
Furthermore, an action-tuning strategy is utilized to adjust the LLM with precision, aligning its predictive capabilities with the specific demands of the long-term action anticipation task. The collective refinement of these elements yields a streamlined approach. Such an approach not only enhances the operational efficiency of LLMs in the context of our study but also provides a foundation for subsequent scholarly inquiry.

Sequential text, as the original input form of the LLM, guides the model in understanding action semantics. The application of LLMs across various multimodal tasks \cite{yeo2024akvsr,aafaq2022dense,gao2024knowledge,chen2024flapping} highlights their substantial representational capabilities. Accordingly, we predict the text category corresponding to preceding frames to effectively utilize the LLM's commonsense knowledge. To accurately capture the distinct feature of both textual and visual modalities while identifying their inherent connections, we develop a Cross-Modality Interaction Block (CMIB). The CMIB is designed for portability and can be applied across a wide range of tasks that require multimodal interaction. It facilitates inter-modal complementarity and verification, particularly in situations where noisy or incomplete unimodal information is encountered. \figref{example} provides an illustration of the ActionLLM framework predicting actions within a kitchen scenario. In cases where the noisy text ``add tea bag'' is input, the model correctly predicts subsequent actions by augmenting it with additional visual information. 

ActionLLM utilizes the pre-trained sequence modeling capabilities of LLMs to process entire sequences at once, bypassing the step-by-step hidden state propagation. Such an approach improves the capture of long-range dependencies and enhances computational efficiency. Additionally, integrating multimodal information bolsters the model's comprehension of intricate scenarios, thereby enhancing its generalization. Such integration effectively alleviates the shortcomings of LSTM and RNN-based methods~\cite{abu2018will,ke2019time,sener2020temporal}.
The proposed CMIB fuses multimodal data, retaining the advantages of the transformer without significantly increasing parameters. Freezing most pre-trained parameters and simplifying the model further enhances parameter efficiency, resolving the inefficiencies that Transformer-based approaches~\cite{gong2022future,nawhal2022rethinking,bhagat2024let} face with long sequences.
Evaluations on the 50 Salads and Breakfast datasets show that multimodal fine-tuning enhances the performance of LLMs in long-term action anticipation.

In summary, our contributions are as follows:
\begin{itemize}
\item We introduce ActionLLM, a pioneering effort in harnessing LLMs for the long-term action anticipation task. ActionLLM effectively leverages the robust sequential modeling and semantic comprehension abilities of LLMs to handle long-term action sequences.
\item We incorporate additional text modalities to tap into the inherent commonsense of LLMs. To explore the unique characteristics of each modality and enable interaction between visual and textual modalities, we design a Cross-Modality Interaction Block.
\item Extensive experiments demonstrate that the proposed method achieves superior performance on benchmark datasets. ActionLLM opens new avenues for long-term action anticipation, encouraging further exploration of the potential of LLMs in enhancing action anticipation.

\end{itemize}

\section{Related Works}
\subsection{Long-term Action Anticipation}
Action anticipation aims to forecast future actions by analyzing preceding actions. The development of large-scale video datasets~\cite{kuehne2014language,sener2022assembly101,grauman2022ego4d} significantly advances research in the field of action anticipation. In contrast to short-term action anticipation~\cite{girase2023latency,furnari2020rolling,osman2021slowfast}, which focuses on immediate future actions, long-term action anticipation involves extended time spans and complex action sequences, placing higher demands on memory capacity and anticipatory reasoning. Early approaches~\cite{abu2018will,ke2019time,sener2020temporal} primarily employ RNN, CNN, and LSTM models to capture dynamic changes in long-term time series, but these methods often suffer from error accumulation and reduced prediction efficiency. Several studies~\cite{abu2021long,moniruzzaman2022jointly} employ cycle consistency to ensure that predicted actions align with observed actions in both semantic and feature spaces. Additionally, some research~\cite{abu2019uncertainty,furnari2018leveraging} highlight that modeling the uncertainty of future actions can be effective for long-term action anticipation.

Recent transformer-based studies~\cite{zhang2024object,patsch2024long,li2022locality} have significantly advanced long-term action anticipation. For instance, Gong \etal~\cite{gong2022future} introduce an end-to-end model that combines global self-attention with parallel decoding. Moreover, Nawhal \etal~\cite{nawhal2022rethinking} utilize both segment-level representations from various activity segments and video-level representations. Additionally, Bhagat \etal~\cite{bhagat2024let} enhance the transformer-based attention mechanism by predicting human intentions from brief observation contexts and incorporating domain knowledge. Current methods for long-term action anticipation primarily rely on either past action labels from segmentation models~\cite{li2023unified,zhou2020motion,ge2021video,chen2025general,liang2023local,zhou2022survey} or visual features from previous frames. In contrast, our study integrates both textual and visual information about past actions, leveraging their complementary strengths. Such an integrated approach captures temporal dependencies and contextual information more effectively, resulting in more accurate and robust predictions. It also mitigates the limitations of relying solely on action labels, which can be error-prone, or visual features, which may lack semantic clarity.

\subsection{Multimodal Learning}
Multimodal learning~\cite{yeo2024akvsr,aafaq2022dense,gao2024knowledge,zhou2020m} handles the complexity of real-world data by providing a more comprehensive and nuanced understanding. Core principles of multimodal learning include modal complementarity and information alignment. Recently, numerous multimodal models~\cite{qi2022simultaneously,xing2023dual,wang2021polo} have been introduced. 
In video understanding, Zhang \etal~\cite{zhang2023video} propose the Video-LLaMA framework, which integrates visual and auditory signals to generate meaningful responses in large models. In addition, Qi \etal~\cite{qi2022simultaneously} introduce a one-phase compression mechanism for Vision-and-Language Pre-training models that streamlines the traditional ``pre-training then compressing'' process. Several studies explore cross-modal interaction to enhance performance in short-term action anticipation, using modalities such as optical flows~\cite{furnari2020rolling,osman2021slowfast} and audio~\cite{zhong2023anticipative}. Multimodal learning is also being explored for long-term action anticipation~\cite{mittal2022learning,zhang2024object}. Zhang \etal~\cite{zhang2024object} leverage object-centric representations and pretrained visual-language models to improve future action predictions.

Unlike existing multimodal methods for the long-term action anticipation task, our approach incorporates an additional text modality in conjunction with visual modalities. Such integration aims to leverage the commonsense knowledge obtained from LLMs, which enhances the model’s ability to understand and predict complex action sequences. The text modality provides contextual insights and semantic understanding that complement the visual data, offering a richer and more nuanced interpretation of the action dynamics. To facilitate modality integration, we introduce a novel Cross-Modality Interaction Block. The module is designed to enable efficient and seamless interaction between the different modalities, ensuring that the information from text and visual sources is harmoniously combined.

\subsection{Parameter-Efficient Fine-Tuning for LLMs}
With the rapid advancement of LLMs~\cite{chowdhery2023palm,touvron2023llama,li2024lmeye}, parameter-efficient fine-tuning methods~\cite{houlsby2019parameter,li2021prefix,lester2021power} have become a primary focus of research and have garnered widespread attention. Existing parameter-efficient fine-tuning methods~\cite{liu2023gpt,hu2021lora,liu2024dora} are classified into three categories based on where the additional training parameters are integrated: Adapter Tuning~\cite{houlsby2019parameter}, Prefix Tuning~\cite{li2021prefix,liu2023gpt,lester2021power}, and Low-Rank Adaptation~\cite{hu2021lora,liu2024dora,wang2024LLMAction}. Recently, Liu \etal~\cite{liu2024dora}  introduce the Weight-Decomposed Low-Rank Adaptation, which decomposes pre-trained weights into magnitude and direction components, updating only the direction component using LoRA. Many studies~\cite{luo2024cheap, huang2023palm} enhance the efficacy of LLMs on downstream tasks by using PEFT methods. For instance, Song \etal~\cite{song2024pneumollm} introduce a framework for medical image diagnosis utilizing LLMs, focusing on data-scarce diseases such as pneumoconiosis. 
Additionally, Luo \etal~\cite{luo2024cheap} propose the Mixture-of-Modality Adaptation for tuning LLMs for vision-language tasks which uses a dynamic routing algorithm.
Furthermore, Fan \etal~\cite{fan2025navigation} introduces BEVInstructor which integrates Bird’s Eye View data into Multi-Modal Large Language Models to facilitate the generation of instructions.

A few recent studies~\cite{huang2023palm,wang2024LLMAction} have explored the use of LLMs in the action anticipation task. While these works aim to integrate LLMs into long-term action anticipation, they each remain focused on a single modality: the Palm~\cite{huang2023palm} relies on textual data, and the LLMAction~\cite{wang2024LLMAction} is limited to visual inputs.
In contrast, our approach leverages a multimodal framework that fuses both visual and textual information to enhance the semantic understanding of actions. Specifically, ActionLLM introduces a Cross-Modality Interaction Block, which facilitates deeper interaction between the two modalities, resulting in more comprehensive modeling of action sequences. Additionally, ActionLLM includes a streamlined output mechanism and an action-tuning module, which optimizes the LLM's efficiency for long-term action prediction.
Compared to existing methods, our approach is more innovative in its use of multimodal information fusion and cross-modal interaction, offering a more robust solution for complex multimodal scenarios.

\section{Method}
\setcounter{sectionCounter}{3}

In this section, we provide an overview of the proposed method, ActionLLM. Sec.\hyperref[sec:method-overall]{\Roman{section}-A} outlines the framework of ActionLLM, detailing the model's holistic architecture. Subsequent sections offer in-depth descriptions of the methods employed in each component of the model. Sec.\hyperref[sec:method-CMIB]{\Roman{section}-B} introduces the Cross-Modality Interaction Block, which is designed to integrate diverse modalities. Sec.\hyperref[sec:method-text]{\Roman{section}-C} describes refinements in LLMs and the approach for obtaining action labels. Additionally, Sec.\hyperref[sec:method-Traing]{\Roman{section}-D} discusses the strategies and objectives of the training process.

\subsection{Overall Model Architecture}
\label{sec:method-overall}

\begin{figure*}[htbp]
\centerline{\includegraphics[width=\linewidth]{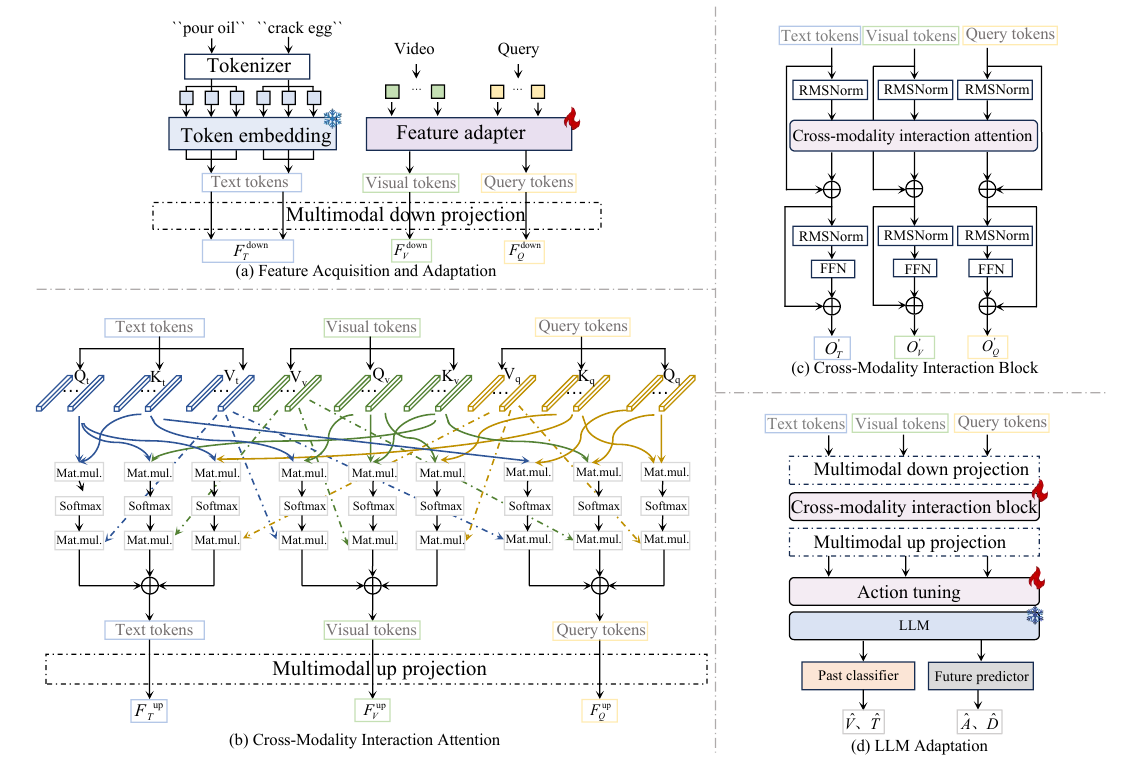}}
\captionsetup{justification=raggedright,singlelinecheck=false}
\caption{Network architecture of ActionLLM. (a) Feature Acquisition and Adaptation. The raw action label is tokenized to generate text tokens, which are then processed through the token embedding layer of the frozen LLM to extract text features. The visual I3D features and the preset query are passed through a feature adapter layer to align with text features. (b) Cross-Modality Interaction Attention (CMIA). CMIA employs self-attention and cross-attention mechanisms to thoroughly investigate the distinct characteristics of each modality and the inter-relationships between them. We use arrows of various shapes and colors to differentiate the flow of data: blue for textual features, green for visual features, and yellow for query processing. Dashed arrows highlight the processing of V in the attention mechanism to clarify the CMIA module's internal structure. (c) Cross-Modality Interaction Block (CMIB). CMIB outputs textual, visual, and action query features following modal fusion. (d) LLM Adaptation. The action tuning module is used to fine-tune the LLM to handle the long-term action anticipation. The multimodal down/up projection layer ensures compatibility with the input specifications of both CMIB and LLM. The outputs from the transformer layers of LLM are processed through their respective past classifier and future predictor to produce future actions.}
\vspace{-0.4cm}
\label{frame_work}
\end{figure*}

For long-term action anticipation, a complete video $V$ with $T$ frames $\left\lbrace{f_{1},f_{2},...,f_{T}}\right\rbrace$ and $I$ action classes $\left\lbrace{A_{1},A_{2},...,A_{I}}\right\rbrace$ is divided into three segments: observation $V_{obs}$ , prediction $V_{pre}$ and remaining $V_{rem}$. The lengths of $V_{obs}$ and $V_{pre}$ depend on observation ratio $\alpha$, prediction ratio $\beta$ and original length of the video ($\alpha+\beta\le 1$). 
We use the I3D feature extractor~\cite{carreira2017quo} on $V_{obs}$ to derive the visual representations $\left\lbrace{f^{'} _{1} ,f^{'} _{2} ,...,f^{'}_{\alpha T}}\right\rbrace$. Simultaneously, we obtain the corresponding action labels $\left\lbrace{a _{1} ,a _{2} ,...,a_{\alpha T}}\right\rbrace$ via a classification model. Although the classifier may introduce some noise into the labels, it provides valuable semantic information. Subsequently, we define $N$ queries $\left\lbrace{q_{1} ,q_{2} ,...,q_{N}}\right\rbrace$ to predict future actions $\left\lbrace{\hat{a}_{1},\hat{a}_{2},...,\hat{a}_{N}}\right\rbrace$ and their respective duration $\left\lbrace{\hat{d}_{1},\hat{d}_{2},...,\hat{d}_{N}}\right\rbrace$, which are used to anticipate the actual future actions $\left\lbrace{\hat{a}_{\alpha T+1},\hat{a}_{\alpha T+2},...,\hat{a}_{(\alpha + \beta)T}}\right\rbrace$.

We use LLMs for the long-term action anticipation task. 
\figref{frame_work} depicts ActionLLM's components and structure, with sub-figures labeled in the order of data processing. Subsequently, we will introduce the functionality of each sub-figure in the sequence of data input.
\figref{frame_work} (a) depicts the process of processing multimodal data.
The text modality is transformed into rich semantic features through word segmentation and embedding. The visual modality leverages the I3D model to extract keyframe features. Concurrently, the preset action queries are combined with the visual features in the feature adapter to produce features that align with the text modality in the same feature space.

We design a Cross-Modality Interaction Block to improve multimodal integration and action understanding.
\figref{frame_work} (c) and \figref{frame_work} (b) present the structure of the Cross-Modality Interaction Block and the internal processing details of cross-modal interaction attention, respectively. During the interaction between visual and textual modalities, we incorporate preset action queries to provide anticipated information about future actions, optimizing and guiding the model's predictions. The Cross-Modality Interaction Block not only plays a crucial role in the long-term action anticipation task but also offers new perspectives and solutions for multimodal data processing.

In \figref{frame_work} (d), the multimodal projections act as a conduit, enabling the seamless integration of multimodal data into the CMIB and the LLM. We preserve the LLM's existing knowledge by freezing its parameters, while applying action tuning to tailor it to the specific needs of the long-term action anticipation task. Simultaneously, we develop a past classifier to process historical text and visual modalities, and a future predictor designed for upcoming actions.
\subsection{Cross-Modality Interaction Block}
\label{sec:method-CMIB}
The sequence modeling capability of the LLM, combined with visual information, enhances the model's reasoning about action causality. Integrating textual information further aids the model to achieve a deeper understanding of the evolution of action patterns. However, achieving effective alignment and fusion of text and visual modalities, to fully exploit the informational potential of both, remains challenging. To address this issue, we design a Cross-Modality Interaction Block (CMIB) to optimize the information integration process and improve the model's multimodal learning capabilities. Cross-modal interaction attention (CMIA) is the core of the CMIB. In \figref{frame_work} (b), we depict the internal structure of CMIA.

To enable the CMIA to focus on the representative features inherent to each modality, we implement the self-attention mechanism across individual modal dimensions. 
For the input features $I_{T}$, $I_{V}$, $I_{Q}$ of the CMIA, the procedure is outlined as follows:
\begin{equation}
MHSA\left ( I \right ) = MHA\left ( I,I \right ) , I=I_{T} ,I_{V},I_{Q},
\end{equation}
\begin{equation}
MHA\left ( X,Y \right ) =\left ( attn_{1}\left ( X,Y \right ),...,attn_{h}\left ( X,Y \right )   \right )  W^{O},
\end{equation} 
\begin{equation}
attn_{i}\left ( X,Y \right ) =\sigma\left ( \frac{\left ( XW_{i}^{Q}   \right )\left ( YW_{i}^{K}   \right )^{T}   }{\sqrt{D/h} }  \right )YW_{i}^{V},   
\end{equation} 
where $attn_{i}\left ( X,Y \right )$ denotes the attention mechanism for inputs $X$,$Y$ with $i$ as the $i$-th head. H is the number of heads, $\sigma$ represents the softmax operation and D is the input dimension. $W_{i}^{T}$, $W_{i}^{V}$, $W_{i}^{Q}$, $W^{O}$ represent linear layers. Multi-head attention results from computing $attn$ on multiple heads. 
When the two different inputs $X$ and $Y$ of the multi-head attention are replaced by the same input $I$, the outcome is multi-head self-attention. In this context, $I$ can be $I_{T}$, $I_{V}$, $I_{Q}$.
The analysis of the text modality reveals implicit semantic connections between action labels, providing crucial contextual clues. In the visual modality, the model identifies features related to character actions by discerning fine-grained details across consecutive frames and filtering out irrelevant background information. The forward-looking self-attention mechanism of the action query enables the model to predict potential consequences and future changes by identifying implicit patterns and trends in the current data.

At the same time, we employ the cross-attention mechanism to facilitate feature fusion. For each modality, we enable interactions both with the other two and within the modality itself. Mathematically, the process is described as follows:
\begin{equation}
O_{T} = MHSA\left ( I_{T} \right )+MHA\left ( I_{T} ,I_{V} \right )+MHA\left ( I_{T} ,I_{Q} \right ),
\end{equation}
\begin{equation}
O_{V} = MHSA\left ( I_{V} \right )+MHA\left ( I_{V} ,I_{T} \right )+MHA\left ( I_{V} ,I_{Q} \right ),
\end{equation}
\begin{equation}
O_{Q} = MHSA\left ( I_{Q} \right )+MHA\left ( I_{Q} ,I_{T} \right )+MHA\left ( I_{Q} ,I_{V} \right ),
\end{equation}
where $O_{T}$, $O_{V}$, $O_{Q}$ signify the output of the text, vision, and query modes of CMIA, respectively. 
The interaction between textual and visual modalities allows the model to leverage information from multiple sources, which improves its ability to reconstruct past action contexts. Such integration helps the model avoid the limitations of relying on a single modality and supports the generation of representations that maintain both semantic consistency and contextual understanding. 

By incorporating the visual modality with action queries, the model is able to detect implicit patterns within visual data that may not be immediately apparent. Additionally, the semantic information from the text modality provides valuable temporal context, which is crucial for understanding the sequence and duration of actions. The temporal dependency enables the model to capture the dynamics of actions over time. When the text modality is combined with action queries, the model’s predictive capability for future actions is enhanced by utilizing the coherent relationships within the textual data, leading to more accurate forecasts of upcoming events.

In the CMIB, the inputs $F_{T}^{down}$, $F_{V}^{down}$, $F_{Q}^{down}$ are combined with the one-dimensional positional encoding. The fused vectors are then normalized by RMSNorm and processed by CMIB, producing outputs $O_{T}$, $O_{V}$, $O_{Q}$. These outputs are further refined with residual connections and an additional application of RMSNorm. Finally, a feed-forward network, along with another residual connection, generates the final output of the CMIB:
\begin{equation}
RMSNorm\left ( r \right ) =\frac{r}{\sqrt{Mean(r^{2}+\varepsilon _{0} )}} W_{R},
\end{equation} 
\begin{equation}
O=CMIB\left ( RMSNorm\left ( F^{down} +P \right )  \right ) +F^{down},
\end{equation}
\begin{equation}
O^{'} =FFN\left ( RMSNorm\left ( O \right )  \right ) +O,
\end{equation}
where $r$ denotes the input to RMSNorm, $\varepsilon _{0}$ stands for a constant, and $W_{R}$ is a learnable scale. $F^{down}$ in the formula can be replaced by $F_{T}^{down}$, $F_{V}^{down}$, $F_{Q}^{down}$, and $O_{T}$, $O_{V}$, $O_{Q}$ can be obtained accordingly. 
Ultimately, the CMIB yields the outputs $O^{'}_{T}$, $O^{'}_{V}$ and $O^{'}_{Q}$. 

\subsection{Refinements in LLM}

\label{sec:method-text}
To fully leverage the potential of LLMs for long-term action anticipation, we introduce several key architectural refinements that enhance their ability to process and integrate multimodal information effectively. These refinements are focused on improving the model’s ability to handle text and visual inputs concurrently while streamlining the prediction process for increased efficiency and accuracy.

First, we expand the LLM’s input capabilities beyond traditional text data to include both visual information and predefined action queries. To ensure that these diverse inputs are aligned within the model, we implement a feature adaptation strategy. The strategy involves mapping visual and query features into a shared feature space that is compatible with the LLM’s text processing pipeline. 
For predictive text acquisition, we use fine-tuned ResNet-50 and ViT-L models to generate text classification labels. Known for their strong performance in image recognition and simple architectures, these models are well-suited for basic classification tasks. We conduct four-fold cross-validation on the Breakfast dataset and five-fold cross-validation on the 50 Salads dataset, training separate models for each fold to ensure reliability. To improve robustness against label inaccuracies, we introduce controlled text noise during action anticipation training.

In addition, we restructure the output mechanism of the LLM to suit the demands of the action anticipation task. Instead of relying on complex regression-based text decoding layers, we replace them with a simple linear layer. The change reduces the computational burden and enhances the model’s efficiency without sacrificing accuracy. Furthermore, we employ a parallel decoding approach for predicting future action sequences, which allows the model to generate predictions quickly and with great consistency.

To further refine the LLM’s predictive capabilities, we introduce an action tuning module. The module fine-tunes the LLM specifically for the task of long-term action anticipation, preserving its inherent commonsense knowledge while optimizing it for the particular challenges posed by multimodal prediction. By freezing the majority of the LLM’s parameters and only adjusting those most relevant to action prediction, we maintain the model’s generalization ability while ensuring it is finely tuned for the specific application.

These refinements collectively enhance the performance of the LLM in multimodal settings, enabling it to accurately predict complex sequences of actions over extended periods. The streamlined architecture not only improves computational efficiency but also makes the model accessible for practical applications, setting a strong foundation for future research in leveraging LLMs for multimodal tasks.

\subsection{Training Strategies and Objectives}
\label{sec:method-Traing}

\subsubsection{Adaptation}

\begin{figure}[t]
\centerline{\includegraphics[width=\linewidth]{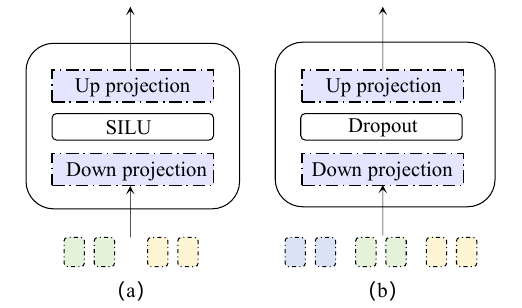}}
\captionsetup{justification=raggedright,singlelinecheck=false}
\caption{Adaptation modules. (a) Feature Adapter. This module is responsible for harmonizing the dimensions and representations between visual and query features. (b) Action Tuning. This module focuses on optimizing and fine-tuning the architecture of LLMs to enhance their performance in specific downstream tasks. }
\vspace{-0.4cm}
\label{adaptation}
\end{figure}

As shown in \figref{adaptation}, we adopt two key tuning strategies: Feature Adapter and Action Tuning. The Feature Adapter aligns the feature spaces of the visual modality and action query with that of the text modality by mapping them to a common vector space, ensuring consistency across these features throughout the analytical process. Action Tuning involves detailed adjustments to LLMs, aiming to preserve their rich prior knowledge while enhancing their accuracy and responsiveness in long-term action anticipation.

Firstly, we sample the past text and visual signals at sampling rate $\mu _{0}$ to obtain $F_{V}=\left [ f^{'}_{1},f^{'}_{2}...,f^{'}_{\theta _{0}  }\right ] ^{T}$ \( \in \mathbb{R}^{\theta _{0}  \times L_{D} }\)  and $A=\left [ a_{1},a_{2}...,a_{\theta _{0}}\right ] ^{T}$ \( \in \mathbb{R}^{\theta _{0}  \times 1 }\), where $ \theta _{0} =\lfloor \frac{\alpha T}{\mu _{0} } \rfloor$, and $\lfloor \cdot \rfloor$ denotes the floor operation. $F_{Q}=\left [ q_{1},q_{2}...,q_{N}\right ] ^{T}$ \( \in \mathbb{R}^{ D  \times L_{D} }\) is determined based on the pre-set query length $N$ and the dimension of the visual feature $L_{D}$. 
$F_{V}$ and $F_{Q}$ are processed with the following formula:
\begin{equation}
F_{V} ^{down}=MD\left [ W_{F_{1} } \left [ SiLU\left ( W_{F_{0}}\left ( F_{V} \right ) +b_{F_{0} }  \right )  \right ]+b_{F_{1} }   \right ], 
\end{equation} 
\begin{equation}
F_{Q} ^{down}=MD\left [ W_{F_{1} } \left [ SiLU\left ( W_{F_{0}}\left ( F_{Q} \right ) +b_{F_{0} }  \right )  \right ]+b_{F_{1} }   \right ], 
\end{equation}
where $W_{F_{0}}$ \( \in \mathbb{R}^{ L_{D} \times L_{M_{F}} }\),$W_{F_{1}}$ \( \in \mathbb{R}^{ L_{M_{F}} \times L_{E} }\) are the weights of the linear layer involved in the feature adapter, $b_{F_{0} }$ and $b_{F_{1} }$ are their bias terms. $L_{D}$,  $L_{M_{F}}$, $L_{E}$ denote the dimensions of the input, middle mapping, and output of the feature adapter. $MD[\cdot]$ represents the multimodal down projection layer.

At the same time, action labels $A=\left [ a_{1},a_{2}...,a_{\theta _{0}}\right ] ^{T}$ \( \in \mathbb{R}^{\theta _{0}  \times 1 }\) are processed by the LLM's default tokenizer, decomposing it into tokens. Then they are converted into text features $F_{T}=\left [ t_{1},t_{2}...,t_{\theta _{0}}\right ] ^{T}$ \( \in \mathbb{R}^{\theta _{0} \times 1 }\) using token embedding. Since a single action label $a_{1}$ can correspond to multiple tokens, we average the embedding vectors to obtain $t_{1}$,  where $t_{1}$ represents the aggregated embedding vector for the action label $a_{1}$. Averaging the vectors ensures the number of text features $F_{T}$ aligns with the number of past observed actions $\theta _{0}$.
\begin{equation}
t_{1}=Mean\left [ Embedding \left ( Tokenizer\left ( a_{1} \right )  \right )  \right ],
\end{equation} 
After $F_{T}$ passes through $MD[\cdot]$, we obtain the inputs $F_{T}^{down}$, $F_{V}^{down}$,and  $F_{Q}^{down}$ for the CMIB. The CMIB explores the intrinsic connections and interactions among the three modes. The results from the CMIB pass through the $MU[\cdot]$, which represents multimodal up projection layer, yielding $F_{T}^{up}$, $F_{V}^{up}$, and $F_{Q}^{up}$.

By concatenating $F_{T}^{up}$, $F_{V}^{up}$, and $F_{Q}^{up}$ through residual connections, we obtain $F_{M}$ \( \in \mathbb{R}^{\left ( 2\theta _{0}+N  \right )  \times L_{E} }\). As the input for action tuning, $F_{M}$ is processed as follows:
\begin{equation}
F_{M} = Cat\left ( F_{T}^{up}+ F_{T}, F_{V}^{up}+ F_{V},F_{Q}^{up}+ F_{Q} \right ), 
\end{equation} 
\begin{equation}
F_{M} = RMSNorm\left ( F_{M} \right ),
\end{equation} 
\begin{equation}
F_{M}= W_{C_{1} } \left [ Dropout\left ( W_{C_{0}}\left ( F_{M} \right ) +b_{C_{0} }  \right )  \right ]+b_{C_{1} }, 
\end{equation} 
where $W_{C_{0}}$ \( \in \mathbb{R}^{ L_{E} \times L_{M_{A}} }\),$W_{C_{1}}$ \( \in \mathbb{R}^{ L_{M_{A}} \times L_{E} }\) are the weights of the convolutional Layer involved in the feature adapter, $b_{C_{0} }$ and $b_{C_{1} }$ are their bias terms. $L_{M_{A}}$ denotes the middle dimension of the action tuning.

\subsubsection{Training Objective}
The training objective is divided into two main components: past and future, each further subdivided into two subtasks. In our loss function, each loss component contributes equally to the total loss.
\begin{equation}
\mathcal{L}_{total}=\mathcal{L}_{V}+\mathcal{L}_{T}+\mathcal{L}_{A}+\mathcal{L}_{D},
\end{equation} 
For the past component, we use visual and textual features to generate two action segmentation losses $\mathcal{L}_{V}$ and $\mathcal{L}_{T}$. Action segmentation involves decomposing a series of actions into distinct units, with each unit representing a specific action. 
\begin{equation}
L_{S}=-\sum_{i=1}^{\theta _{0}}\sum_{j=1}^{K}S_{(i,j)} log \hat{S}_{(i,j)}, 
\end{equation} 
where $\theta _{0}$ represents the number of past actions, $K$ denotes the types of future actions, $S_{(i,j)}$ indicates the labels of past actions, and $\hat{S}_{(i,j)}$ represents the logits. Replacing $\hat{S}_{(i,j)}$ with either $\hat{V}_{(i,j)}$ or $\hat{T}_{(i,j)}$ yields results $L_{V}$ or $L_{T}$.

For the future component, we calculate the loss for future action categories, $\mathcal{L}_{A}$ and their duration loss $\mathcal{L}_{D}$ based on the action queries. 
\begin{equation}
L_{A}=-\sum_{i=1}^{N}\sum_{j=1}^{K+1}A_{(i,j)} log \hat{A}_{(i,j)} \mathbb{I}_{i \leq \varphi },
\end{equation} 
\begin{equation}
L_{D}=-\sum_{i=1}^{N}\left ( D_{i}-\hat{D_{i}}    \right ) ^{2} \mathbb{I}_{i < \varphi }.
\end{equation} 
where $A_{(i,j)}$ and $D_{i}$ are labels for the categories of future actions and their durations, respectively. In the indicator function $\mathbb{I}_{i < \varphi }$, $\varphi $ refers to the position of the action class ``None''.
According to FUTR~\cite{gong2022future}, the predicted actions include an additional class labeled ``None'' compared to the GT, which indicates when the model should stop predictions.

\section{Experiments}

\begin{table*}[htbp]
  \centering
  \caption{Comparison of the state-of-the-art methods addressing long term action anticipation on the 50 Salads. Each column represents a distinct $\alpha$-$\beta$ combination, with the optimal mean over classes (\%) highlighted in bold and the second-best underlined.}
    \begin{tabular*}{\textwidth}{@{\extracolsep{\fill}}>{\centering\arraybackslash}m{1.5cm}|>{\centering\arraybackslash}m{2.5cm}|*{4}{>{\centering\arraybackslash}m{1cm}}|*{4}{>{\centering\arraybackslash}m{1cm}}|>{\centering\arraybackslash}m{1.5cm}}
    \hline
    \multirow{1}[4]{*}{Inputs} & \multirow{1}[4]{*}{Works} & \multicolumn{4}{c|}{$\beta$($\alpha$ = 0.2)} & \multicolumn{4}{c|}{$\beta$($\alpha$ = 0.3)} & \multirow{1}[4]{*}{Average} \\
\cline{3-10}          &       & 0.1 & 0.2 & 0.3 &0.5 & 0.1 & 0.2 &  0.3 & 0.5 &  \\
    \hline
    \multirow{4}[2]{*}{Label} & RNN~\cite{abu2018will}   & 30.06  & 25.43  & 18.74  & 13.49  & 30.77  & 17.19  & 14.79  & 9.77  & 20.03  \\
          & CNN~\cite{abu2018will}   & 21.24  & 19.03  & 15.98  & 9.87  & 29.14  & 20.14  & 17.46  & 10.86  & 17.97  \\
          & UAAA~\cite{abu2019uncertainty}  & 24.86  & 22.37  & 19.88  & 12.82  & 29.10  & 20.50  & 15.28  & 12.31  & 19.64  \\
          & Time-Cond~\cite{ke2019time} & 32.51  & 27.61  & 21.26  & 15.99  & 35.12  & 27.05  & 22.05  & 15.59  & 24.65  \\
    \cline{1-11}
    \multirow{10}[4]{*}{Features} & Tem-Agg~\cite{sener2020temporal} & 25.50  & 19.90  & 18.20  & 15.10  & 30.60  & 22.50  & 19.10  & 11.20  & 20.26  \\
          & Cycle Cons~\cite{abu2021long} & 34.76  & 28.41  & 21.82  & 15.25  & 34.39  & 23.70  & 18.95  & 15.89  & 24.15  \\
          & J-AAN~\cite{moniruzzaman2022jointly} & -     & -     & -     & -     & 34.90  & 25.80  & \textbf{24.40}  & 16.10  & - \\
          & ObjectPrompt~\cite{zhang2024object} & 37.40  & 28.90  & 24.20  & 18.10  & 28.00  & 24.00  & \underline{24.30}  & \textbf{19.30}  & 25.53  \\
          & FUTR~\cite{gong2022future}  & 39.55  & 27.54  & 23.31  & 17.77  & 35.15  & 24.86  & 24.22  & 15.26  & 25.96  \\
          & LLMAction~\cite{wang2024LLMAction} & 32.30  & 28.75  & 22.94  & 18.52  & \underline{37.97}  & 24.34  & 24.27  & 15.02  & 25.51  \\
          & Con-Alig.~\cite{patsch2024long} & \underline{41.10}  & \underline{31.40}  & \underline{24.80}  & \underline{19.20}  & 35.80  & \underline{27.50}  & 23.30  & 18.10  & \underline{27.65}  \\
\cline{2-11}  &\cellcolor{lightgray}  Ours  &\cellcolor{lightgray}\textbf{43.67}     &\cellcolor{lightgray}\textbf{32.80}       &\cellcolor{lightgray}\textbf{25.73}       &\cellcolor{lightgray}\textbf{19.51}       &\cellcolor{lightgray}\textbf{44.95}       &\cellcolor{lightgray}\textbf{29.06}       &\cellcolor{lightgray}23.15       &\cellcolor{lightgray}\underline{18.37}       &\cellcolor{lightgray}\textbf{29.66}  \\
    \hline
    \end{tabular*}%
  \label{50 salads}%
\end{table*}%

\begin{table*}[htbp]
  \centering
  \caption{Comparison of the state-of-the-art methods addressing long term action anticipation on the Breakfast. Each column represents a distinct $\alpha$-$\beta$ combination, with the optimal mean over classes (\%) highlighted in bold and the second-best underlined.}
    \begin{tabular*}{\textwidth}{@{\extracolsep{\fill}}>{\centering\arraybackslash}m{1.5cm}|>{\centering\arraybackslash}m{2.5cm}|*{4}{>{\centering\arraybackslash}m{1cm}}|*{4}{>{\centering\arraybackslash}m{1cm}}|>{\centering\arraybackslash}m{1.5cm}}
    \hline
    \multirow{1}[4]{*}{Inputs} & \multirow{1}[4]{*}{Models} & \multicolumn{4}{c|}{$\beta$($\alpha$ = 0.2)} & \multicolumn{4}{c|}{$\beta$($\alpha$ = 0.3)} & \multirow{1}[4]{*}{Average} \\
\cline{3-10}          &       & 0.1 & 0.2 & 0.3 & 0.5 & 0.1 & 0.2 & 0.3 & 0.5  \\
    \hline
    \multirow{3}[2]{*}{Label} & RNN~\cite{abu2018will}   & 18.11  & 17.20  & 15.94  & 15.81  & 21.64  & 20.02  & 19.73  & 19.21  & 18.46  \\
          & CNN~\cite{abu2018will}   & 17.90  & 16.35  & 15.37  & 14.54  & 22.44  & 20.12  & 19.69  & 18.76  & 18.15  \\
          & UAAA~\cite{abu2019uncertainty}  & 16.71  & 15.40  & 14.47  & 14.20  & 20.73  & 18.27  & 18.42  & 16.86  & 16.88  \\
          & Time-Cond~\cite{ke2019time} & 18.41  & 17.21  & 16.42  & 15.84  & 22.75  & 20.44  & 19.64  & 19.75  & 18.81  \\
    \hline
    \multirow{4}[4]{*}{Features} & CNN~\cite{abu2018will}   & 12.78  & 11.62  & 11.21  & 10.27  & 17.72  & 16.87  & 15.48  & 14.09  & 13.76  \\
          & Tem-Agg~\cite{sener2020temporal} & 24.20  & 21.10  & 20.00  & 18.10  & \underline{30.40}  & 26.30  & 23.80  & 21.20  & 23.14  \\
          & Cycle Cons~\cite{abu2021long} & \underline{25.88}  & \textbf{23.42}  & \textbf{22.42}  & \textbf{21.54}  & 29.66  & \underline{27.37}  & \underline{25.58}  & \textbf{25.20}  & \underline{25.13}  \\
\cline{2-11}  &\cellcolor{lightgray}  Ours  &\cellcolor{lightgray}\textbf{26.38}       &\cellcolor{lightgray}\underline{23.32}       &\cellcolor{lightgray}\underline{21.70}       &\cellcolor{lightgray}\underline{20.43}       &\cellcolor{lightgray}\textbf{31.77}       &\cellcolor{lightgray}\textbf{29.45}       &\cellcolor{lightgray}\textbf{26.68}       &\cellcolor{lightgray}\underline{24.55}       &\cellcolor{lightgray}\textbf{25.54}  \\
    \hline
    \end{tabular*}%
  \label{breakfast}%
\end{table*}%

\subsection{Experiment Setups}

\subsubsection{Datasets}
The Breakfast dataset~\cite{kuehne2014language} consists of 1,712 videos featuring 52 individuals engaged in multiple breakfast preparation activities across 18 different kitchen settings. Each video averaging approximately 2.3 minutes in length, is annotated with 48 distinct fine-grained action labels. The videos, captured at a frame rate of 15 fps and downsampled to a resolution of 240×320 pixels, provide a comprehensive view of the breakfast-making process. Following prior research~\cite{abu2018will,abu2019uncertainty,gong2022future}, the 4-fold cross-validation is implemented for the Breakfast assessment.

The 50 Salads~\cite{stein2013combining} dataset encompasses 50 top-view videos featuring 25 individuals engaged in the task of salad preparation. The dataset spans over 4 hours of RGB-D video footage, captured at a resolution of 640×480 pixels and a frame rate of 30 fps. Each video averages around 6.4 minutes and encompasses approximately 20 distinct action instances. These actions are annotated with 17 fine-grained labels, as well as 3 high-level activity categories. Following prior research~\cite{abu2018will,abu2019uncertainty,gong2022future}, the dataset includes a standard protocol of 5-fold cross-validation, with average results reported across all splits.

\subsubsection{Quantitative Metric}
Following prior research~\cite{patsch2024long,moniruzzaman2022jointly}, we use mean over classes accuracy (MoC) as the quantitative metric, calculated as the average accuracy of all future time points within a specified anticipation duration. As the protocol outlined by Farha \etal~\cite{abu2018will}, we set observation ratio $\alpha =\left [ 0.2,0.3 \right ] $ and prediction ratio $\beta =\left [ 0.1,0.2,0.3,0.5 \right ]$.

\subsubsection{Implementation Details}
Among various open-source models~\cite{javaheripi2023phi,almazrouei2023falcon}, LLaMA~\cite{touvron2023llama} stands out for its effectiveness and feasibility, specifically the LLaMA-7B model, making it our choice for the foundational LLM.
Due to the 50 Salads dataset containing less data, fewer action classes, and longer videos compared to the Breakfast dataset, we apply specific processing methods to each dataset following prior research~\cite{abu2018will,abu2019uncertainty,gong2022future}. Both datasets are trained for 20 epochs. And the learning rate is set at 1e-4 for the Breakfast dataset and 1e-3 for the 50 Salads dataset during training. The observation ratio $\alpha$ is adjusted to 20\%, 30\%, and 50\% for the Breakfast dataset and 10\%, 20\%, 30\%, 40\%, and 50\% for the 50 Salads dataset. The sampling rate is set to 6 for the Breakfast dataset and 8 for the 50 Salads dataset. To balance the dataset sizes during training, we set a fixed start frame at 0 for Breakfast and a variable start frame from 0 to 7 for 50 Salads.
In the inference phase, the prediction ratio $\beta$ is consistently set to 50\%, with variations derived from this base. The start frame for both datasets is set to 0. We first use a fine-tuned ResNet-50~\cite{he2016deep} or ViT-L~\cite{dosovitskiy2020image} to classify the observed frames and generate corresponding textual information, which is processed by the LLM to extract textual features. These features are then combined with I3D visual features and fed into the trained ActionLLM for action anticipation under different settings.

Following Luo \etal~\cite{luo2024cheap}, 16-bit quantization is applied to the LLM to augment video memory efficiency. The output of the LLM is refined by substituting the intricate text decoding layer with a linear layer. Predicted texts undergo processing through the model's tokenizer and word embedding layer. These enhancements facilitate the efficient execution of experiments on an NVIDIA RTX 3090 Ti.

\subsection{Comparision with State-of-the-Art Works}
\tabref{50 salads} presents a comparison of ActionLLM with other leading methods on the 50 Salads dataset, emphasizing different $\alpha$-$\beta$ configurations. Among the previous best performers, the method such as ``Con-Alig''~\cite{abu2021long} demonstrates notable success by aligning contextual features, achieving a mean accuracy of 27.65\%. However, ActionLLM pushes these boundaries further, particularly at an $\alpha$ of 0.3 and $\beta$ of 0.1, where it achieves a remarkable accuracy of 44.95\%. 
The improvement can be attributed to the advanced sequence modeling capabilities of LLMs, which are further enhanced by our fine-tuning strategies and the incorporation of multimodal data. 
Furthermore, the results demonstrate that our approach, by integrating visual and textual data, effectively overcomes the limitations of relying solely on a single modality. Specifically, compared to LLMAction, which only uses visual information, ActionLLM shows a significant performance improvement.

In \tabref{breakfast}, we compare the ActionLLM framework against several methods on the Breakfast dataset. Notably, the Transformer-based methods, such as "Cycle Consistency" [1], have previously set the benchmark by effectively capturing temporal dependencies over long sequences, achieving a mean accuracy of 25.13\% across different $\alpha$-$\beta$ settings. ActionLLM surpasses the performance of "Cycle Consistency", particularly excelling in the scenario with an $\alpha$ of 0.3 and a $\beta$ of 0.2, where we achieve a mean accuracy of 29.45\%. 
This phenomenon demonstrates the superior sequential modeling capabilities of LLMs, which leverage extensive pre-trained commonsense knowledge to anticipate long-term actions more accurately.
Compared to Con-Alig~\cite{patsch2024long}, which uses contextual alignment to capture dependencies between actions, and J-AAN~\cite{moniruzzaman2022jointly}, which employs self-knowledge distillation and cycle consistency for the same purpose, ActionLLM leverages the inherent sequential modeling capabilities of LLMs to model the dependencies.
Furthermore, ActionLLM requires fewer trainable parameters than the Transformer-based approach presented by Gong et al.~\cite{gong2022future}.
The incorporation of CMIB allows for the fusion of visual and textual information, leading to enhanced predictive performance, especially in complex scenarios where conventional approaches fall short.

\begin{table}[htbp]
  \centering
  \caption{Comparison of learnable parameters, frozen parameters, FLOPs and inference time (M: million, B: billion, G: GigaFLOPs, s: second) across different methods, including FUTR, LLMAction, and ActionLLM.}
    \begin{tabular}{c|cc|c|r}
    \hline
    \multirow{1}[3]{*}{Method} & \multicolumn{2}{c|}{Parameters} & \multirow{1}[3]{*}{FLOPs} & \multicolumn{1}{c}{\multirow{1}[3]{*}{Time}} \\
\cline{2-3}          & Learnable & Frozen &       &  \\
    \hline
    FUTR  & 17.38M   & 0     & 7.07G & 0.07s \\
    LLMAction & 3.90M    & 7B    & 5574.96G & 1.41s \\
    \rowcolor{lightgray}ActionLLM & 4.21M    & 7B    & 5575.18G & 1.36s  \\
    \hline
    \end{tabular}%
  \label{flops}%
\end{table}%

\tabref{flops} indicates that while LLM-based methods do not excel in total parameter count, they have fewer trainable parameters than FUTR and perform better with limited data. Reducing the volume of parameters is a potential future research direction. We compare the FLOPs of the FUTR, LLMAction, and ActionLLM, ensuring a fair comparison by maintaining a consistent input sequence length of 423 and fixing the dimension size at 2048 for the three models. ActionLLM, with its multimodal features, increases trainable parameters and computational load but is versatile across various multimodal scenarios.

In \tabref{flops}, we also compare the inference times of three models on a single NVIDIA RTX 3090 Ti. To ensure a fair comparison of inference time, we test three models on a same video data with a fixed observation ratio of 0.2 and a prediction ratio of 0.3. Due to differences in prediction methods, the input sizes vary: ActionLLM, LLMAction, and FUTR have input sequence lengths of 624, 1588, and 423, respectively, with input dimensions of 4096, 4096, and 512. Additionally, inference time is affected by hardware, software configurations, and other uncertainties.
Although the LLM-based methods, ActionLLM and LLMAction, exhibit longer inference times primarily due to larger parameter sizes, they demonstrate robust temporal mining capabilities. Moving forward, our future work will explore lightweight versions of LLMs to retain their strong temporal analysis capabilities while improving speed.

ActionLLM outperforms on the 50 Salads dataset compared to the Breakfast dataset due to the latter's higher complexity, with 2.5 times more action categories. Such complexity complicates text label prediction for past frames, resulting in lower accuracy on the Breakfast dataset and impacting final predictions. Accurate past text labels are crucial in multimodal prediction, as they provide essential semantic guidance for LLMs, ensuring reliable outcomes.

\subsection{Ablation Study}

\subsubsection{The Effectiveness of CMIB in Enhancing Multimodal Integration}
As shown in \tabref{is multi}, CMIB outperforms both CMIB-S, which relies solely on self-attention, and CMIB-C, which utilizes only cross-attention. The results indicate that neither CMIB-S nor CMIB-C alone is sufficient for multimodal interaction, underscoring the effectiveness of the full CMIB that combines both mechanisms. To further validate CMIB's contribution, we also test the configuration without CMIB, referred to as CMIB-N, which feeds multimodal features directly into the LLM. The results demonstrate a significant performance drop with CMIB-N, confirming that CMIB plays a crucial role in leveraging the complementary strengths of visual and textual modalities.

\begin{table}[t]
  \centering
  \caption{CMIB Ablation. We compared the MoC(\%) of CMIB with CMIB-S, which uses only self-attention, and CMIB-C, which uses only cross-attention, and further removed the CMIB module to validate its effectiveness (CMIB-N).}
    \begin{tabular}{c|cccc|r}
    \hline
    \multicolumn{1}{c|}{\multirow{1}[4]{*}{Method}} & \multicolumn{4}{c|}{$\beta$($\alpha$ = 0.3)}  & \multicolumn{1}{c}{\multirow{1}[4]{*}{Avg. }} \\
\cline{2-5}          & 0.1 & 0.2 & 0.3 & 0.5 &  \\
    \hline
    CMIB-S     &36.87       &26.27       &22.76       &14.51       &25.10  \\
    CMIB-C     &40.43       &25.39      &21.22       &18.42       &26.37  \\
    CMIB-N     &39.61       &25.49       &19.45       &16.02       &25.14  \\
    \rowcolor{lightgray} CMIB     &44.95       &29.06       &23.15       &18.37      &28.88  \\
    \hline
    \end{tabular}%
  \label{is multi}%
\end{table}%

\subsubsection{Assessing the Contribution of Loss Components}
Our loss function consists of four key components: textual action segmentation loss, visual action segmentation loss, future action category loss, and future action duration loss. The results, summarized in \tabref{loss}, demonstrate that each component plays a crucial role in improving the model's accuracy. 
Starting with only the future action losses (category and duration), the model shows reasonable performance. However, when the past visual action segmentation loss is added, we observe a significant improvement, highlighting the importance of capturing visual cues from past frames to enhance future predictions. Adding the textual action segmentation loss further boosts performance, emphasizing that accurate textual information from past frames provides essential semantic context that strengthens the model’s understanding of action sequences.
With all four loss components, the model achieves superior performance due to a balanced learning strategy that integrates both past and future text and visual information.

\begin{table}[htbp]
  \centering
  \caption{Loss Ablation. Through three strategic configurations, we assess the distinct contribution of each loss component. }
  \begin{tabularx}{\linewidth}{
    >{\centering\arraybackslash}X 
    >{\centering\arraybackslash}X 
    >{\centering\arraybackslash}X 
    >{\centering\arraybackslash}X|
    >{\centering\arraybackslash}c 
    >{\centering\arraybackslash}c 
    >{\centering\arraybackslash}c 
    >{\centering\arraybackslash}c|
    >{\centering\arraybackslash}r
  } 
    \hline
    \multicolumn{4}{c|}{Loss} & \multicolumn{4}{c|}{$\beta$($\alpha$ = 0.3)} & \multicolumn{1}{c}{\multirow{1}[3]{*}{Avg.}} \\
    \cline{1-8}
    $\mathcal{L}_{T}$ & $\mathcal{L}_{V}$ & $\mathcal{L}_{A}$ & $\mathcal{L}_{D}$ & 0.1 & 0.2 & 0.3 & 0.5 &  \\
    \hline
    - & - & \checkmark{} & \checkmark{} & 32.64 & 19.51 & 15.85 & 11.30 & 19.83 \\
    - & \checkmark{} & \checkmark{} & \checkmark{} &38.61 &24.62 &21.14 &15.82 &25.05  \\
    \rowcolor{lightgray}\checkmark{} & \checkmark{} & \checkmark{} & \checkmark{} & 44.95 & 29.06 & 23.15 & 18.37 & 28.88 \\
    \hline
  \end{tabularx}
  \label{loss}
\end{table}
\subsubsection{Optimizing Dimensions in Action Tuning and CMIB}
\tabref{query dim} shows that action prediction accuracy increases with hidden layer dimensions up to 4, beyond which it declines. This trend suggests that a dimension of 4 optimally balances underfitting and overfitting. Dimensions lower than 4 (e.g., 2) may miss fine-grained features, while higher dimensions (e.g., 8) can add complexity that hinders generalization.

\tabref{CMIB dim} shows a positive correlation between the dimensions of the CMIB and action prediction accuracy under certain conditions. We choose a 128-dimensional CMIB over a 256-dimensional one to balance performance and computational efficiency, despite the latter's better performance in a specific setting ($\alpha$ = 0.3). The 256-dimensional model has 1.67 times more trainable parameters, resulting in higher computational costs and longer training times. Moreover, when $\alpha$ = 0.2, the 128-dimensional model outperforms the 256-dimensional one, exhibiting superior overall average performance.

\begin{table}[htbp]
  \centering
  \caption{Dimensions for action tuning. We present the performance on the 50 Salads dataset, along with the corresponding number of trainable parameters (in millions) for various dimensions. }
    \begin{tabular}{c|cccc|c|c}
    \hline
    \multirow{1}[4]{*}{Dim} & \multicolumn{4}{c|}{$\beta$($\alpha$ = 0.3)} & \multicolumn{1}{c|}{\multirow{1}[4]{*}{Avg. }} & \multirow{1}[4]{*}{Para.} \\
\cline{2-5}          & 0.1   & 0.2   & 0.3   & 0.5   &       &  \\
    \hline
    2     &37.22       &24.71       &22.40       &16.63      &25.24       &3.68  \\
    \rowcolor{lightgray}4     &44.95       &29.06       &23.15       &18.37      &28.88       &4.21  \\
    6     &40.48       &24.42       &22.13       &18.12      &26.29       &4.73  \\
    8     &35.27       &24.25       &20.33       &17.51       &24.34       &5.25  \\
    \hline
    \end{tabular}%
  \label{query dim}%
\end{table}%

\begin{table}[htbp]
  \centering
  \caption{Dimensions for CMIB. We present the performance on the 50 Salads dataset, along with the corresponding number of trainable parameters (in millions) for various dimensions.}
    \begin{tabular}{c|cccc|c|c}
    \hline
    \multirow{1}[4]{*}{Dim} & \multicolumn{4}{c|}{$\beta$($\alpha$ = 0.3)} & \multicolumn{1}{c|}{\multirow{1}[4]{*}{Avg. }} & \multirow{1}[4]{*}{Para.} \\
\cline{2-5}          & 0.1   & 0.2   & 0.3   & 0.5   &       &  \\
    \hline
    32     &37.95       &25.57       &22.45      &20.02      &26.52       &2.72  \\
    64     &39.94       &24.79       &24.25       &19.45      &27.11       &3.09  \\
    \rowcolor{lightgray}128     &44.95       &29.06       &23.15       &18.37      &28.88       &4.21  \\
    256     &47.34       &29.95       &21.68       &18.04       &29.25      &7.03  \\
    \hline
    \end{tabular}%
  \label{CMIB dim}%
\end{table}%

\subsubsection{Determining Optimal Number of Action Queries}
\begin{figure}[t]
\centerline{\includegraphics[width=\linewidth]{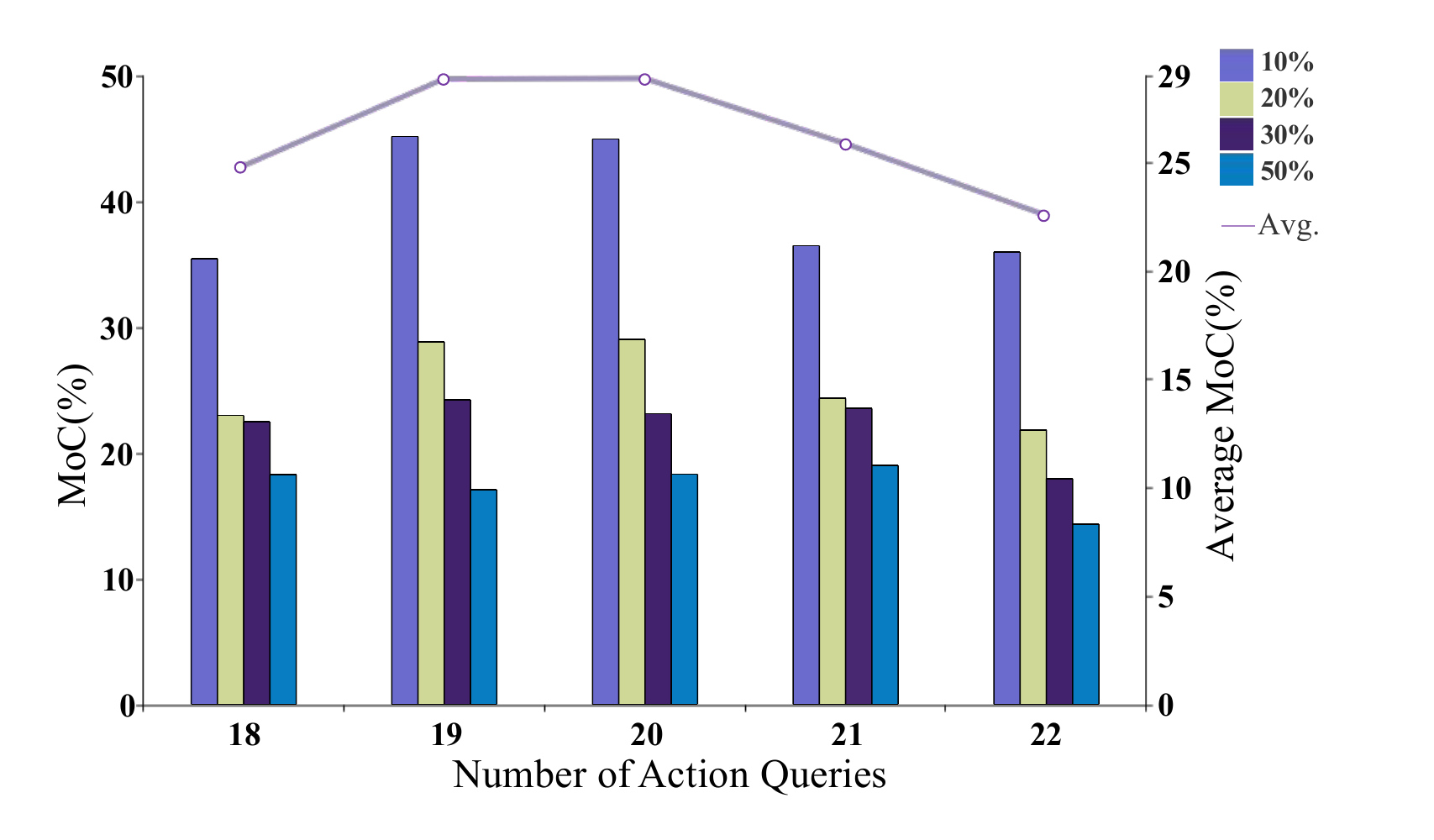}}
\captionsetup{justification=raggedright,singlelinecheck=false}
\caption{We increase the number of action queries until performance degrades. For each query count, we use four prediction ratios (10\%, 20\%, 30\%, and 50\%), keeping the observation ratio fixed at 30\%. Different colors indicate MoC values at each prediction ratio, and the line shows the average MoC change across queries.}
\vspace{-0.4cm}
\label{num_query}
\end{figure}

We perform a grid search to systematically adjust the number of action queries between 18 and 22, aiming to identify the optimal quantity for various datasets. The optimal number is influenced by factors such as the total number of action categories, video length, and action density—the frequency of actions per unit of time. Higher action density may necessitate more action queries to cover the prediction space effectively. As shown in \figref{num_query}, an analysis of the 50 Salads dataset reveals that the most suitable number of action queries for the 50 Salads is 20.

\begin{figure*}[t]
\centerline{\includegraphics[width=\linewidth]{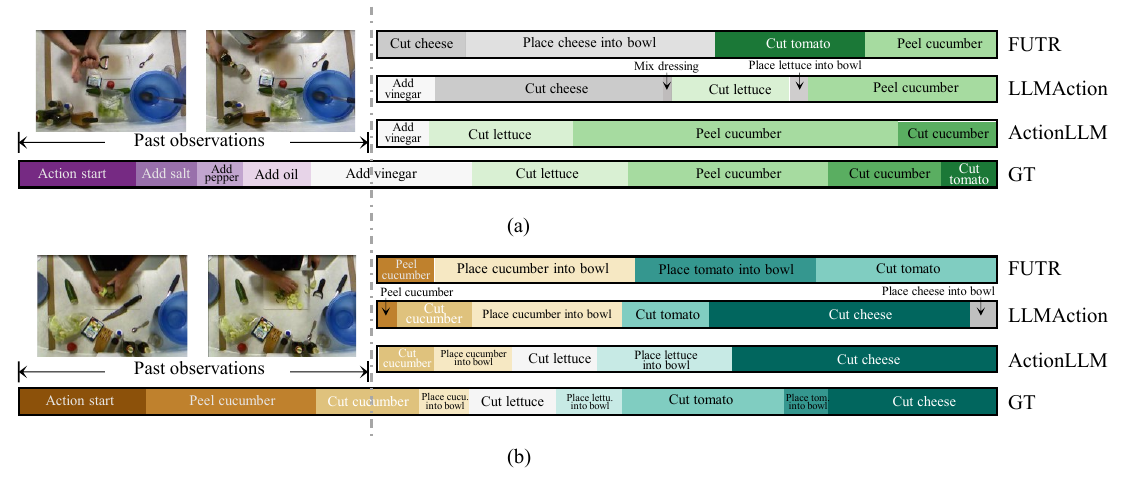}}
\captionsetup{justification=raggedright,singlelinecheck=false}
\caption{Qualitative analysis. We conduct predictions on the future actions of two video examples, labeled (a) and (b), from the 50 Salads dataset using three distinct models: FUTR, LLMAction, and ActionLLM. These predictions are made under experimental settings with $\alpha$ = 0.2 and $\beta$ = 0.3. The transitions between past and future actions are indicated by dashed lines, with each color in each video representing a unique action.}
\vspace{-0.4cm}
\label{vis}
\end{figure*}

\subsubsection{Effect of Action Query Initialization}
From \tabref{query init}, it is evident that the superior performance of ActionLLM when initialized with a constant value, as opposed to zero or random normal initialization, can be attributed to several factors. First, non-zero initialization avoids gradient vanishing or gradient explosion. Constant initialization ensures that weights are not initialized to zero, which is critical for preventing the vanishing or explosion of gradients during the early stages of training. Second, the constant value provides a stable starting point for the weights, facilitating more stable and efficient convergence during optimization. Third, when the constant initialization is properly scaled according to the expected input and output of the network layers, it helps maintain an appropriate activation scale, which is advantageous for training dynamics. And empirical results suggest that the constant initialization method optimally balances the search space of the model's parameters, leading to improved generalization and convergence.
\begin{table}[htbp]
  \centering
  \caption{Query Initialization Methods. An Empirical Evaluation of Zero, Constant, and Random Normal Initializations}
    \begin{tabular}{c|cccc|c}
    \hline
    \multirow{1}[4]{*}{Q} & \multicolumn{4}{c|}{$\beta$($\alpha$ = 0.3)} & \multirow{1}[4]{*}{Avg.} \\
\cline{2-5}          & 0.1   & 0.2   & 0.3   & 0.5   &  \\
    \hline
    0                        &38.34       &29.70       &21.24       &15.11       &26.10  \\
    \rowcolor{lightgray}c    &44.95       &29.06       &23.15       &18.37       &28.88  \\
    randn                    &33.17       &23.56       &22.90       &20.03       &24.92  \\
    \hline
    \end{tabular}%
  \label{query init}%
\end{table}%

\subsection{Qualitative Insights into ActionLLM}

The proposed ActionLLM is qualitatively compared with FUTR and LLMAction, as shown in \figref{vis}. In complex scenarios with many action categories and long prediction durations, ActionLLM's predictions align most closely with the ground truth. For instance, in \figref{vis} (a), while both ActionLLM and LLMAction accurately predict the subsequent ``cut'' action following the ``add vinegar'' action, LLMAction incorrectly identifies the object being cut as ``cheese''. The difference indicates that ActionLLM is more precise in capturing the fine-grained features of the actions.
As depicted in \figref{vis} (b), ActionLLM maintains the logical sequence of actions by performing the ``cut'' action before the ``place'' action, whereas FUTR incorrectly reverses such order. This phenomenon demonstrates ActionLLM's capability to comprehend the intrinsic meaning of actions and their underlying logical relationships.

The primary goal of this study is to validate the ability of LLMs to handle the long-term action anticipation task. For this purpose, we focuse on using LLaMA-7B as the primary model for our experiments. Thus, we do not explore other LLMs, such as those with different architectures or larger scales, which is a limitation of this work. We plan to extend our investigation to encompass a broader range of LLMs in future research to achieve a more exhaustive assessment.

\section{Conclusion}
This paper presents ActionLLM, a pioneering framework that harnesses the capabilities of LLMs for long-term action anticipation. By treating video sequences as sequential tokens and integrating both visual and textual modalities through the Cross-Modality Interaction Block, ActionLLM effectively captures and leverages temporal dependencies, setting a new standard in the field. Our approach demonstrates that LLMs, traditionally used for language tasks, can be adapted to complex multimodal scenarios, offering a robust solution for predicting extended sequences of actions.The experimental validation on benchmark datasets highlights the superior performance of ActionLLM. The results underscore the framework's ability to maintain high accuracy even in challenging prediction settings, illustrating the potential of LLMs in handling intricate temporal dynamics.
Beyond the immediate contributions, this work opens up several avenues for future research. The integration of LLMs in the action anticipation task encourages further exploration into their application for modeling long-term dependencies across various domains. Future studies could build upon this foundation, refining multimodal fusion techniques and exploring other potential applications of LLMs in sequential prediction tasks.

\bibliographystyle{IEEEtran}

\bibliography{egbib}

\begin{IEEEbiography}
[{\includegraphics[width=1in,height=1.25in,clip,keepaspectratio]{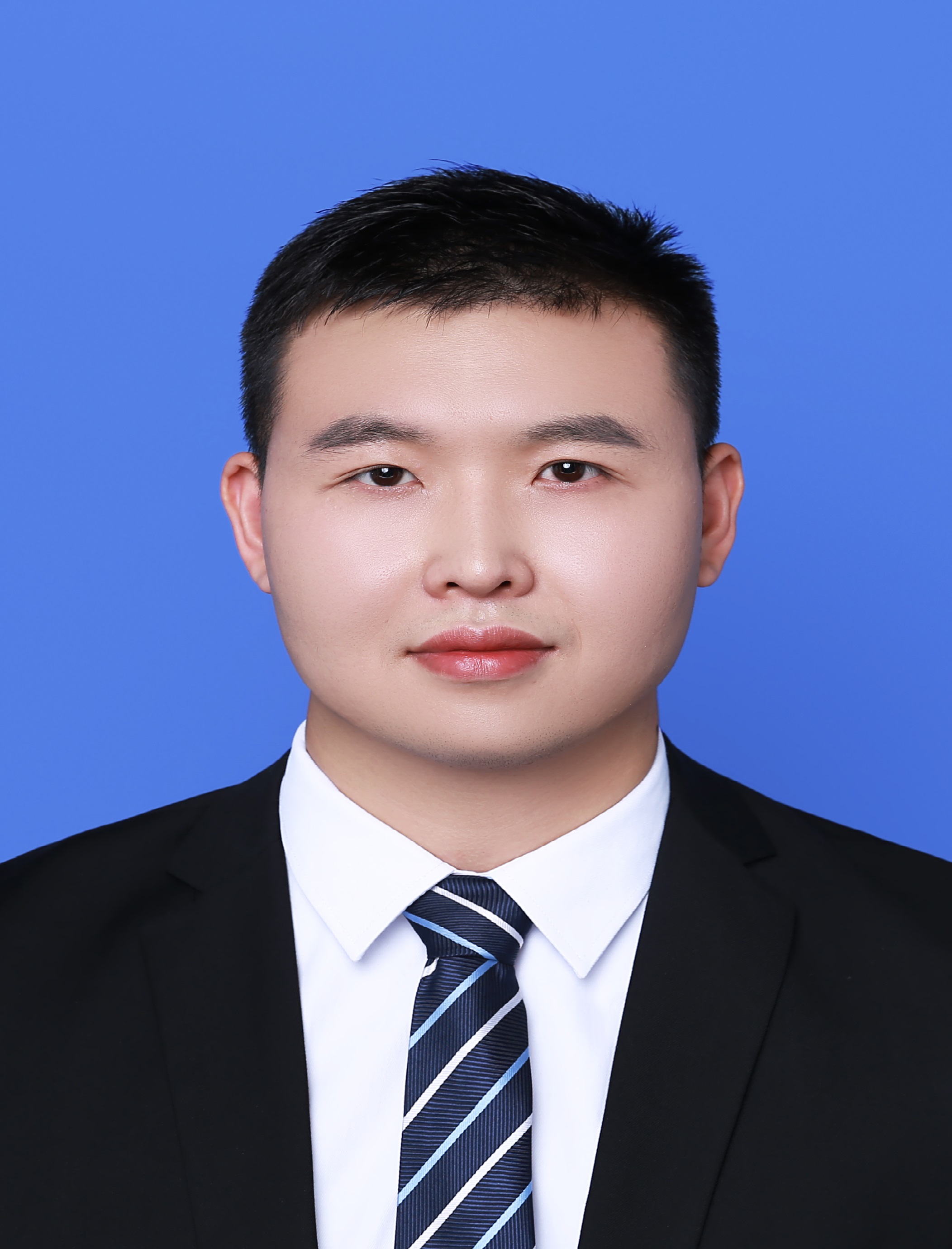}}]
{Binglu Wang} (M'21) received the Ph.D. degree in Control Science and Engineering from the School of Automation at Northwestern Polytechnic University, Xi'an, China, in 2021. He was a postdoctoral researcher with the School of Information and Electronics, Beijing Institute of Technology, Beijing, China, from 2022 to 2024. He is currently a professor with the School of Astronautics, Northwestern Polytechnical University and an adjunct professor with Xi'an University of Architecture and Technology. His research interests include Computer Vision, Digital Signal Processing, and Deep Learning.
\end{IEEEbiography}

\begin{IEEEbiography}
[{\includegraphics[width=1in,height=1.25in,clip,keepaspectratio]{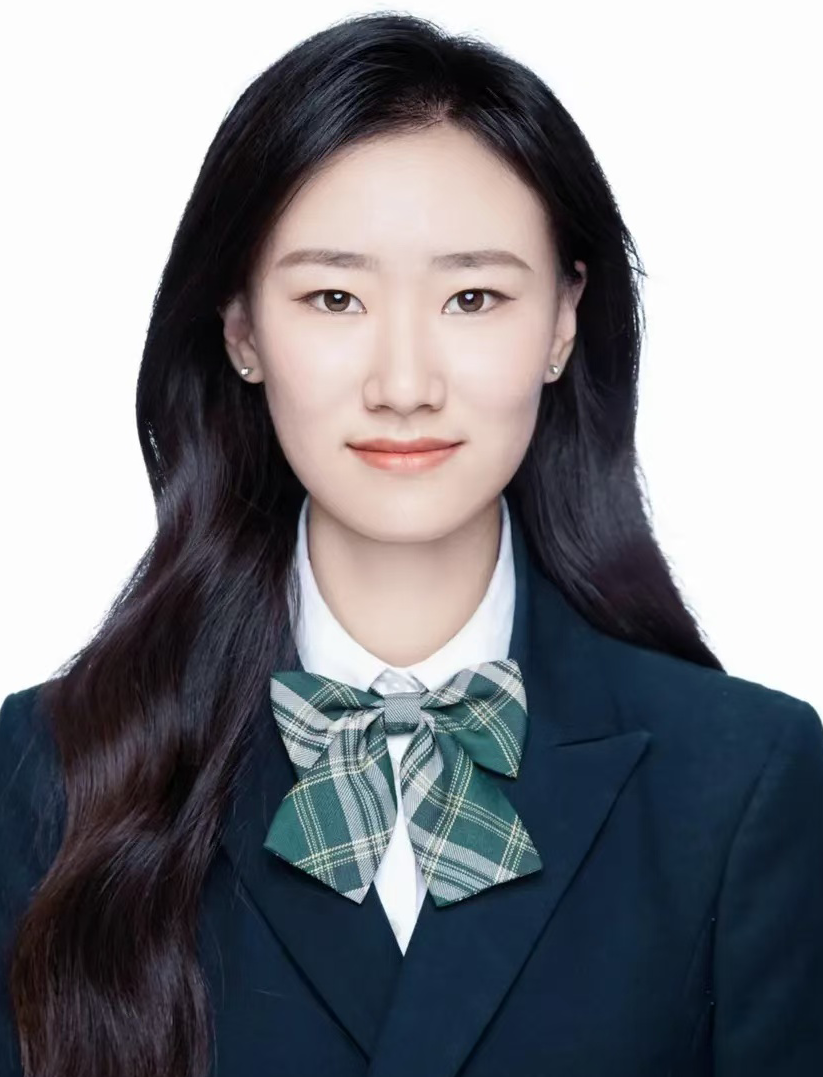}}]
{Yao Tian} received her B.S. degree in 2023, from Xi'an University of Architecture and Technology, Xi'an, China. She is currently a Master student with Xi'an University of Architecture and Technology, Xi'an, China. Her research interests include video analysis and temporal action detection.
\end{IEEEbiography}

\vspace{-30pt}
\begin{IEEEbiography}
[{\includegraphics[width=1in,height=1.25in,clip,keepaspectratio]{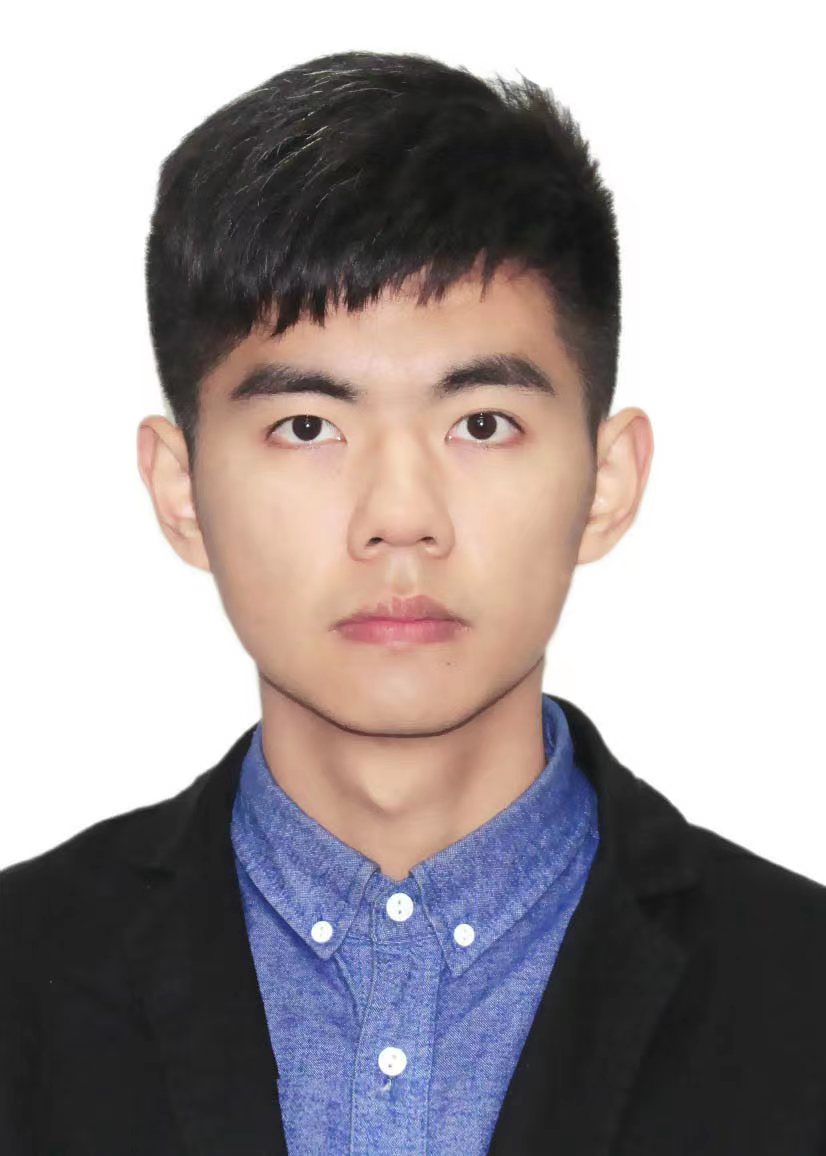}}]
{Shunzhou Wang} (M'24) received the Ph.D. degree from the Beijing Institute of Technology, Beijing, China, in 2023. His research interests include image super-resolution, light field image processing, and video analysis.
\end{IEEEbiography}

\vspace{-30pt}

\begin{IEEEbiography}
[{\includegraphics[width=1in,height=1.25in,clip,keepaspectratio]{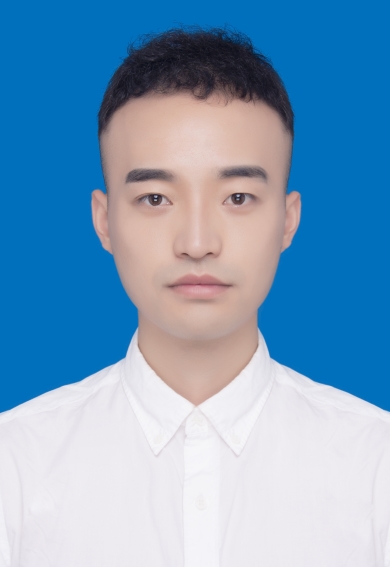}}]
{Le Yang} is currently a post-doctor with Institute of Artificial Intelligence, Hefei Comprehensive National Science Center. He is also with the School of Information Science and Technology, University of Science and Technology of China. His research interests include video analysis and temporal action detection.
\end{IEEEbiography}

\vspace{-30pt}

\end{document}